\pdfoutput=1
\documentclass{bmvc2k}

\usepackage{times}
\usepackage{epsfig}
\usepackage{amsmath}
\usepackage{amssymb}
\usepackage{textgreek}
\usepackage{mathtools}
\usepackage{graphicx}
\usepackage{standalone}
\usepackage{multirow}
\usepackage{booktabs}

\definecolor{darkgreen}{rgb}{0.0, 0.5, 0.05}

\title{Controllable Data Augmentation Through Deep Relighting}

\addauthor{George Chogovadze}{g.tchogovadze@gmail.com}{1}
\addauthor{Rémi Pautrat}{remi.pautrat@inf.ethz.ch}{1}
\addauthor{Prof. Dr. Marc Pollefeys}{marc.pollefeys@inf.ethz.ch}{1}

\addinstitution{
 Computer Vision Group\\
 ETHZ\\
 Zürich, Switzerland\\
}

\runninghead{Chogovadze, Pautrat, Pollefeys}{Augmentation through relighting}

\begin{document}

\maketitle

\begin{abstract}
At the heart of the success of deep learning is the quality of the data. Through data augmentation, one can train models with better generalization capabilities and thus achieve greater results in their field of interest. In this work, we explore how to augment a varied set of image datasets through relighting so as to improve the ability of existing models to be invariant to illumination changes, namely for learned descriptors. We develop a tool, based on an encoder-decoder network, that is able to quickly generate multiple variations of the illumination of various input scenes whilst also allowing the user to define parameters such as the angle of incidence and intensity. We demonstrate that by training models on datasets that have been augmented with our pipeline, it is possible to achieve higher performance on localization benchmarks.
\end{abstract}

\vspace{-0.4cm}
\section{Introduction}
\label{sec:intro}
\vspace{-0.2cm}
With the advent of greater computing power and the accumulation of big data, deep learning has sprung to the forefront of many scientific and engineering disciplines. By solving issues that seemed to require intuitive understanding from a machine, deep neural networks have been outperforming one another year after year. Computer vision has seen great leaps since AlexNet \cite{alexnet} first demonstrated the vast potential that CNNs carry for the field. As newer and deeper architectures are developed, the need for high quality annotated data keeps increasing. Data augmentation bridges the gap by spawning larger sample sets from available datasets. Traditional augmentation techniques have long been used to drastically increase the robustness and the generalization capabilities of deep learning models ~\cite{zhong2017random, taylor2017improving, albumentation, dataaugmentationstudy}. However, the need for more and more specific transformations have led to the development of deep networks with the sole purpose of data augmentation ~\cite{melekhov2020image, DBLP:journals/corr/abs-1909-01056, GAN_Aug, Lemley_2017, MixupGANTraining}. With industry applications of computer vision growing in areas such as autonomous driving, the need for robust models, capable of faring in challenging environments, has increased.

In the aforementioned case, reliably identifying the contents of a scene irrespective of the lighting conditions have proven to be difficult ~\cite{HPatches, VisualLoc}. This is partially due to the inherent difficulties of creating a dataset that would allow the training of an illumination invariant model. Indeed, to collect such a dataset one would not only need to have static contents but also the ability to influence the lighting conditions in a controlled manner. Outdoor samples are a prime example of this challenge. Firstly, the contents of the scene keep evolving chaotically as pedestrians, vehicles and the sky keep changing. Secondly, the lighting provided by the ambient illumination cannot be manipulated. Thus, it is difficult to train a model to have specific characteristics with regards to its robustness to illumination. With the recent improvements in the quality of image generation through generative adversarial networks with specific stylistic properties, it has become possible to influence only the illumination of a scene without affecting the underlying contents.

Hence, we ask ourselves in what manner could image relighting be used as a data augmentation technique and what impact it could have in creating illumination invariant models.
In this work, we introduce a neural network that is able to efficiently relight an image with high fidelity. Based on an encoder-decoder architecture, our model is able to disentangle the illumination of a scene from its geometry. Consequently, it allows the manipulation and replacement of the illumination component, while keeping fine geometrical details. Our illumination encoding is based on the notion of light probes and is therefore easily interpretable and can be manipulated or generated from scratch with modern image software editors. Departing from the style transfer approach, our approach allows the generation of new illuminations without borrowing it from another image. Applied to different models trained on our relit images, we show an increase robustness to illumination changes on several benchmarks. In summary, our work offers the following contributions:
\begin{itemize}
    \item We train an efficient encoder-decoder relighting network on a varied set of scenes by disentangling lighting from geometry.
    \item We provide a lighting encoding based on synthetic light probes and we explore different methods to generate them.
    \item Our lighting data augmentation tool is extensively tested on several benchmarks for local feature detection and description and shows improvement over previous data augmentation tools.
\end{itemize}


\vspace{-0.6cm}
\section{Related Work}
\vspace{-0.1cm}
\subsection{Data Augmentation}
Collecting data is a crucial step in training a model in machine learning. Since the model weights are derived from datapoints, increasing the quantity of variation within a dataset is necessary for a model to converge to its optimal performance. However, acquiring a larger dataset can be very expensive due to the difficulties of labeling \cite{survey} or the simple lack of availability such as in medical imaging. Data augmentation techniques aim to alleviate this problem by introducing artificial samples created from the original existing datasets. Such augmentations are readily available for spatial applications and have been shown to increase the relevant metrics quite substantially~\cite{zhong2017random, taylor2017improving, albumentation, dataaugmentationstudy}. Lately, deep networks have been designed solely for data augmentation tasks~\cite{melekhov2020image, DBLP:journals/corr/abs-1909-01056, GAN_Aug, Lemley_2017, MixupGANTraining}. For example, \citet{dataaugmentationlabel} successfully achieve state of the art performance by simply interpolating scenes between frames of a video. \citet{melekhov2020image} are able to improve on the robustness of existing feature extractors by simply augmenting the training dataset with generated images. Similarly, \citet{GAN_Aug} have concluded that samples created with GANs can compensate for the lack of available data in medical imaging. Our assumption is that by producing synthetically relit images we could achieve similarly promising results.

\vspace{-0.2cm}
\subsection{Image Relighting}

Image relighting can be used for data augmentation as new samples can be generated from an existing image, keeping the same geometry whilst having a different illumination. As it is difficult to have a controlled environment where the underlying geometry of a scene does not change, illumination variants are ideally suited for data augmentation.
The task of image relighting has been studied in several works, both in the literature of deep learning and classical computer vision techniques~\cite{chen2020rnr,Yu2020SelfRelight,helou2020aim,Hou_2021_CVPR,Srinivasan_2021_CVPR,Nestmeyer_2020_CVPR}. Very realistic relighting can be achieved in specific scenarios such as headshot relighting~\cite{relighting,Zhou_2019_ICCV}. Furthermore, various works on style transfer ~\cite{pix2pix, style2huang, style, Perceptual} showed that it is possible to transfer attributes from one scene to another whilst keeping the original contents. Similarly, in the case of city relighting \cite{cityfactor}, it is possible to quantify lighting conditions and transfer them from one image to another. 

The aforementioned works, building on deep learning CNN models, achieve their results in a very specific field with the main focus being on realism and not on the potential applications of data augmentation. Furthermore, each inference model relies on a very specific visual map information for relighting or have very loose control over the qualitative properties of the transferred or modified attributes. By virtue of accounting for occlusions and shadows that would otherwise be difficult to model, higher-level control and better fidelity can be achieved when combining with depth data \cite{DeepLight}.

\vspace{-0.4cm}
\subsection{Feature Disentanglement} 
\vspace{-0.1cm}
 To gain insights on how deep networks actually manipulate data, feature disentanglement has become a highly researched topic in the field of deep learning~\cite{Esser_2020_CVPR,locatello2019,Deng_2020_CVPR,Alharbi_2020_CVPR,harsh2018,Ding_2020_CVPR,awiszus2019,zhu2020,peebles2020,shu2020}. Disentanglement, could not only bring insights about the underlying data but could also allow a finer control on the output of the network. On simple datasets such as MNIST~\cite{mnist_vae,lecun1998mnist} variational autoencoders (VAE) \cite{VAE} are able to achieve competitive disentanglement. However, more complex datasets require stronger constraints to achieve interpretable feature separation. More recently, by imposing constraints on predictable outputs, Hu et al. managed to disentangle features \cite{mixing} by mixing different latent space tensors.  
More commonly, feature disentanglement can be achieved by putting constraints on the orthogonality of the features in the latent space. In their work on \textbeta-VAE \cite{beta-vae}, Higgins et al. demonstrate how a simple penalty term in the loss function can achieve highly disentangled features.
Furthermore, adversarial autoencoder can create defined and controllable features \cite{GANVAE} by imposing constraints on the distribution that variational autoencoders map the latent space to. Finally, \citet{chen2016infogan} demonstrated how to inject disentangled features in the latent space by adding an auxiliary network to the discriminator that would verify if the data had been used.

\vspace{-0.5cm}
\section{Relighting Network}
\begin{figure}[h!]
    \centering
    \includegraphics[width=\linewidth]{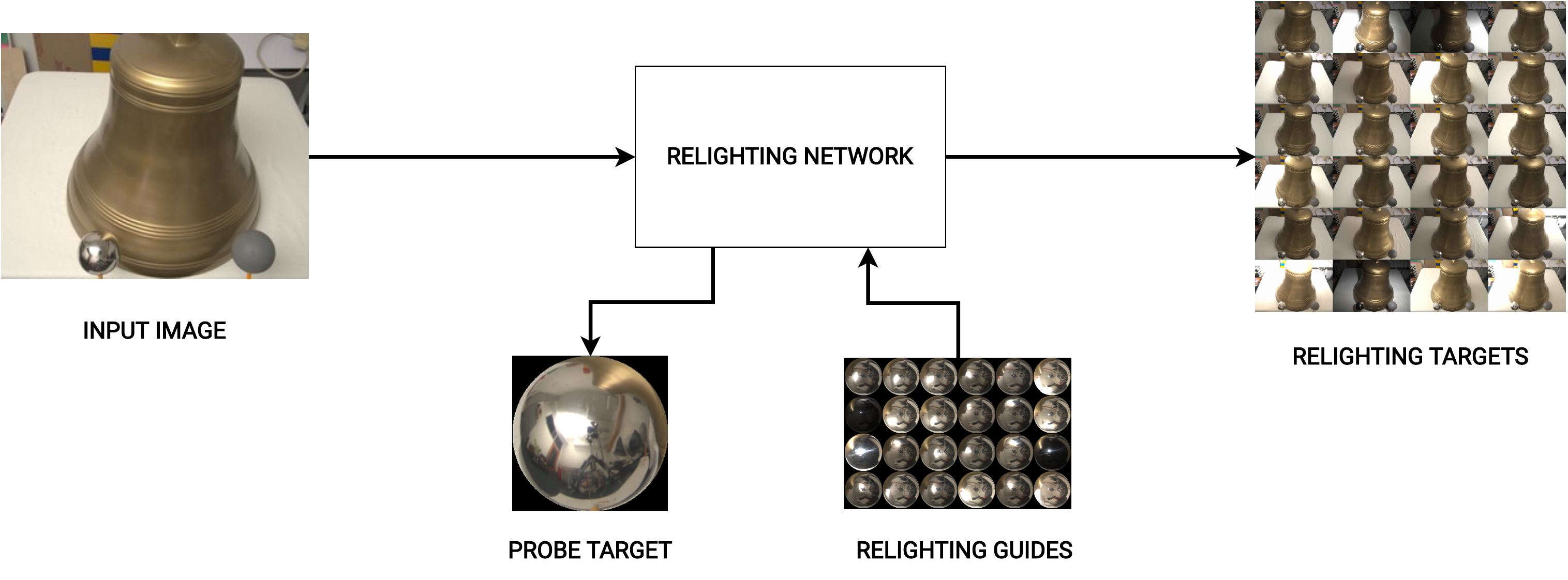}
    
    \vspace{-0.45cm}
    
    \caption{\textbf{Overview of our method.} The network has to predict and generate the original lighting probe. Then by using the input image as geometric reference as well as one of the 24 relighting guides, it has to generate a relit image combining information from both sources.}
     \vspace{-0.4cm}
    \label{fig:training_loop}

\end{figure}

\vspace{-0.2cm}
We introduce an efficient encoder-decoder relighting network based on previous works and insights~\cite{relighting,Zhou_2019_ICCV,Perceptual,UPPROJ,elu}. Starting from a single input RGB image, our network encodes the illumination of the scene in an interpretable lighting encoding in the bottleneck and leverages skip connections to convey the geometrical information of the scene to the newly generated image. Disentanglement is effectively achieved during training by replacing the lighting encoding with a new illumination and asking the network to decode the input scene under the new illumination, as depicted in Figure~\ref{fig:training_loop}.

The strength of our method lies in the lighting encoding, from which one can decode an interpretable light probe displaying the illumination of a scene. This allows a direct interpretation of the lighting predicted by the network and allows us to manipulate the illumination by varying the light direction and intensity for instance (see Section~\ref{sec:inference}).

 \vspace{-0.3cm}
\subsection{Architecture}
\begin{figure}[h!]
\vspace{-0.2cm}
\begin{center}
    \includegraphics[width=\linewidth]{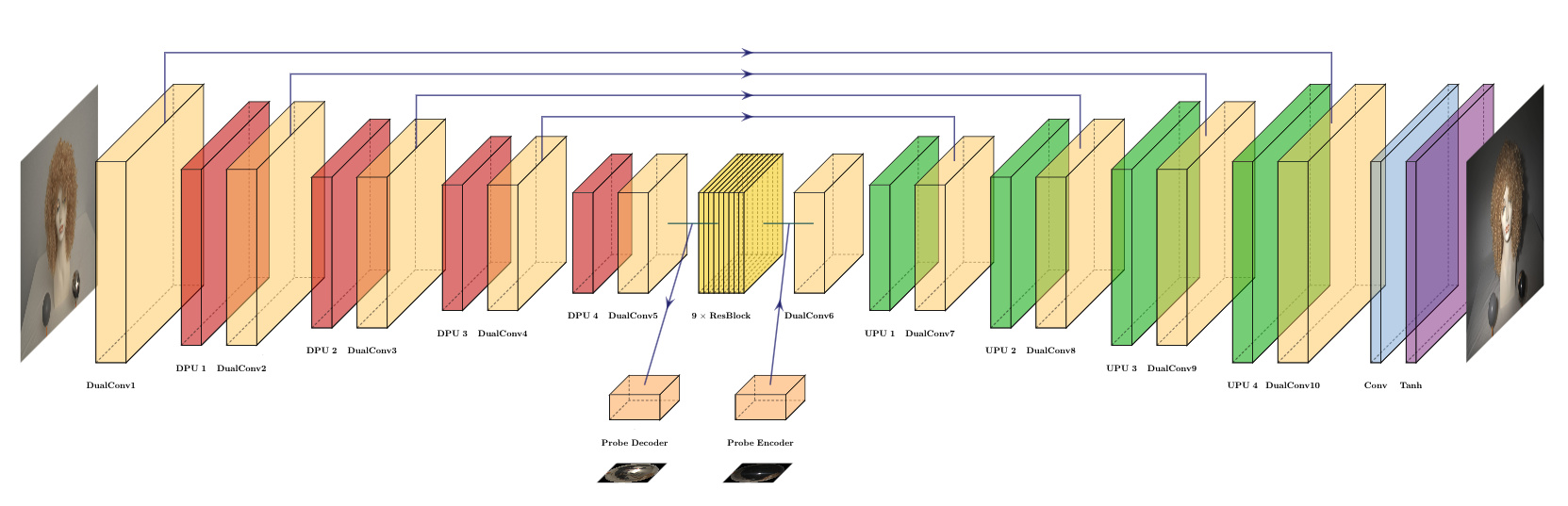}
\end{center}
 \vspace{-0.4cm}
    \caption{\textbf{Our relighting architecture.} We are able to achieve high fidelity image relighting by combining DualConv blocks with projection blocks for sampling~\cite{UPPROJ} and a series of ResBlocks at the bottleneck for better feature extraction similar to Sun et al. \cite{relighting}.}
     \vspace{-0.2cm}
    \label{fig:Detailed Network}

\end{figure}

 \vspace{-0.1cm}
The general architecture is inspired by \citet{relighting}. To alleviate issues with regards to dying neurons and vanishing gradients, we incorporated Up-Projection and Down-Projection blocks \cite{UPPROJ}, coupled with ResBlocks \cite{resblock} and ELU activation functions \cite{elu}. Figure \ref{fig:Detailed Network} summarizes the final network used during training and inference.

    \vspace{-0.5cm}
\subsection{Loss Function}
Our network takes as input an image $I$, predicts the corresponding light probe $\hat{P}$ and using a new lighting probe $P_i$ as guide, it generates a new image $\hat{I_i}$ with the given illumination i corresponding to a ground truth image $I_i$. Denoting $P$ the ground truth light probe of the original image $I$, our total loss is a combination of losses over the predicted light probe and the decoded image. We use an L1 loss to compare the probes and images, and add a perceptual loss~\cite{Perceptual} to increase the fidelity of our generated images:
\begin{align*} 
\tag{1}
L_{\mathrm{probe}} &= \Vert P -  \hat{P} \Vert_1 \\ 
\tag{2}
L_{\mathrm{image}} &= \Vert I_i -  \hat{I_i} \Vert_1 +  \sum_{j=0}^n \Vert l_{j}(I_i) - l_{j}(\hat{I_i})\Vert_2 \\
\tag{3}
L_{\mathrm{total}} &= L_{\mathrm{probes}} + L_{\mathrm{image}}
\end{align*}
The sum in equation number $(2)$ corresponds to the perceptual loss. For each encoding layer $l_{j}$ of a pre-trained VGG network~\cite{simonyan2014deep}, we compute the L2 distance between the encoding of the ground truth and the generated image.
\vspace{-0.25cm}
\subsection{Training}
\vspace{-0.15cm}
We train our network on the Multi-Illumination Dataset (MID)\cite{MID} picturing more than 1000 scenes under 24 lighting conditions. As the light probes available in the dataset are scene-dependent but always correspond to one of the 24 pre-defined illuminations, we create scene-agnostic probes by averaging all the probes corresponding to the same lighting. This effectively creates 24 scene-agnostic probes (displayed on the left of Figure~\ref{fig:syntheticProbes}), which are then used during training. The training lasts for 45 epochs with approximately 10'000 relighting samples per epoch. The total training time is 48 hours on a single 8GB RTX 2070s NVIDIA GPU. The batch size is set to 1 and we use instance normalization at each layer. The learning rate is set to 2e-4 with a scheduler that reduces on plateau by a factor of 1e-1. Each training sample is scaled to a dimension of 256x256.
 \vspace{-0.25cm}
\subsection{Inference}
 \vspace{-0.15cm}
\label{sec:inference}

\begin{figure}[!h]
  \centering
   \hfill
  \begin{minipage}[h]{0.4\textwidth}
    \includegraphics[width=\textwidth]{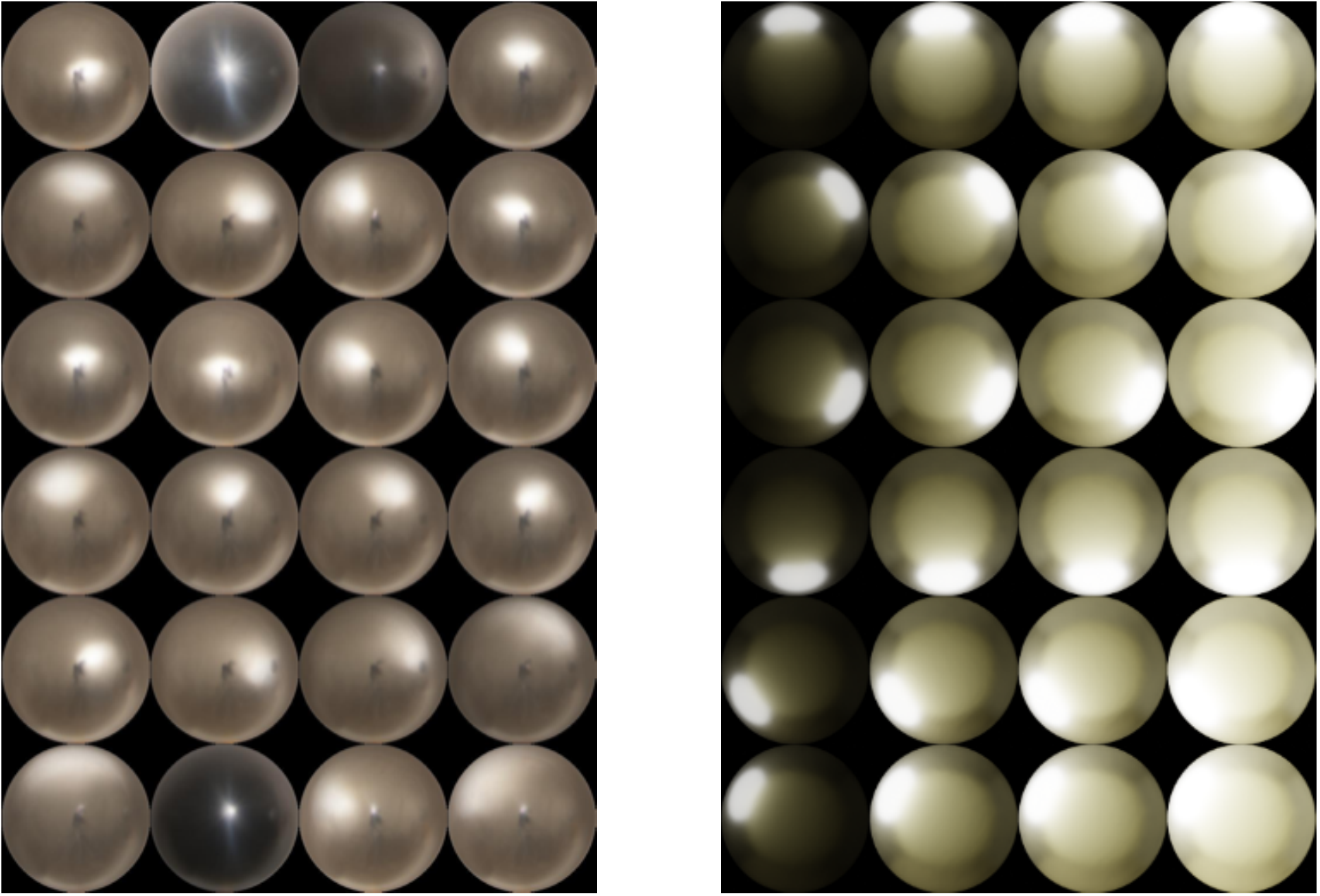}
  \end{minipage}
  \hfill
  \begin{minipage}[h]{0.39\textwidth}
    \includegraphics[width=\textwidth]{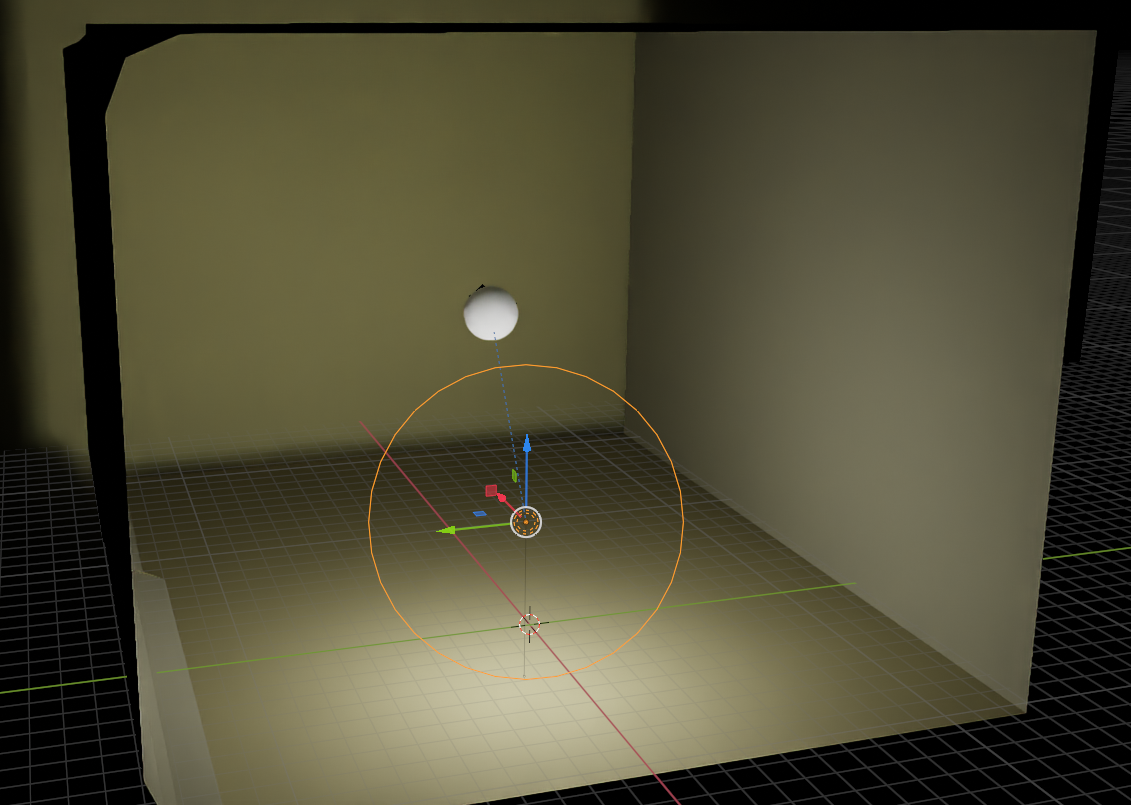}
  \end{minipage}
        \hfill
    \caption{\textbf{Left}: 24 averaged light probes from the MID (left) compared to a few samples of the possible continuous lighting conditions that can be achieved through Blender (middle). \textbf{Right}: Recreated room in Blender to generate new synthetic probes.}
     \vspace{-0.3cm}
\label{fig:syntheticProbes}

\end{figure}

While it is possible to influence the lighting that our network applies to the base scene through the probes that are available in the MID, the control is very limited and lacks granularity. It is difficult to separate things such as the intensity from the angle of incidence. We decided to explore how our model would fare with synthetic probes generated with Blender in an artificial room. Working with Blender allowed us to control the aforementioned lighting properties separately. On Figure \ref{fig:syntheticProbes}, we compare the 24 available directions in the MID, averaged from each scene, to 24 synthetically generated probes from Blender.

\begin{figure}
\begin{minipage}[c]{\textwidth}
\hfill
  \begin{minipage}[c]{0.4\textwidth}
    \includegraphics[width=\textwidth]{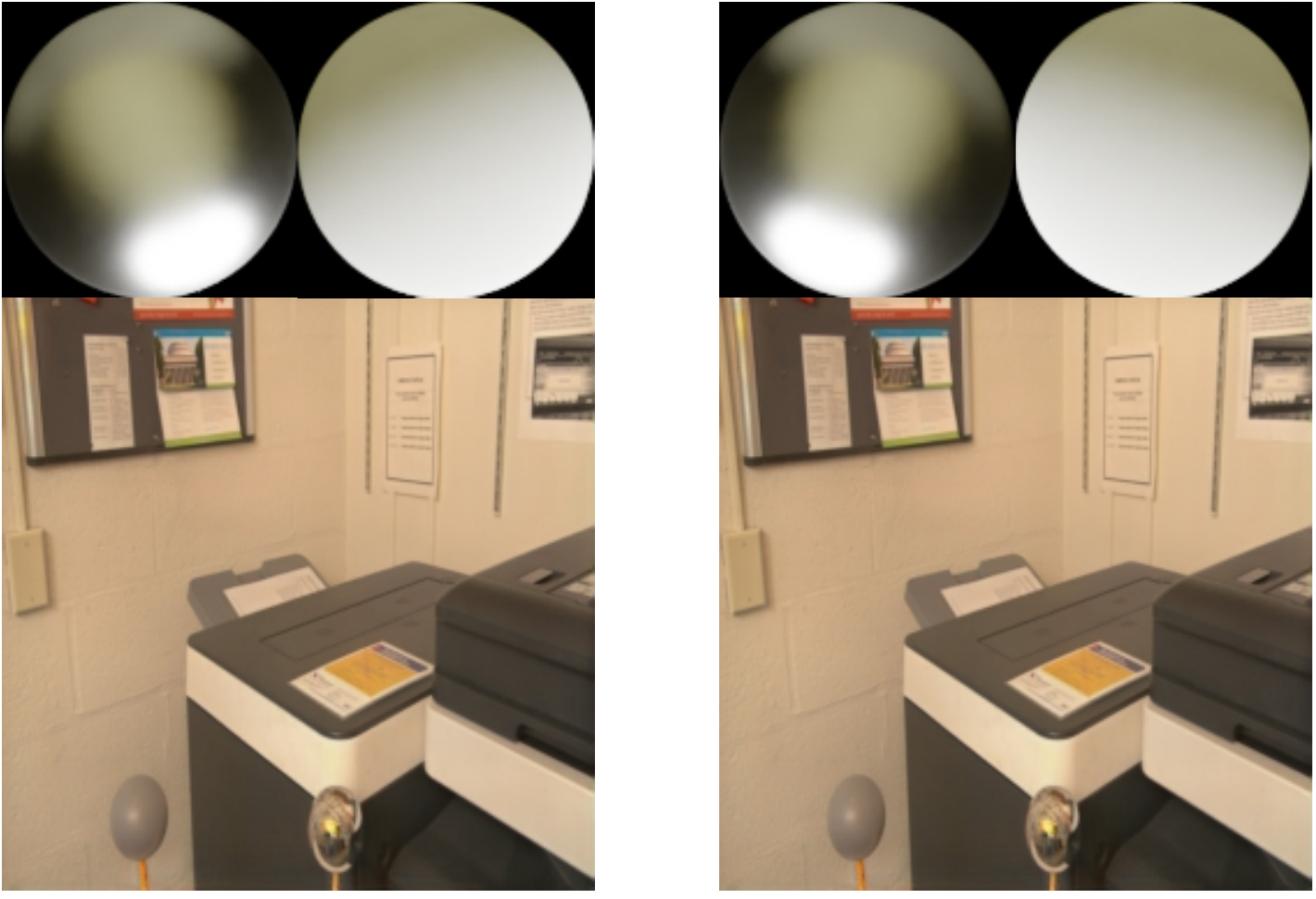}
  \end{minipage}
  \hfill
  \begin{minipage}[c]{0.55\textwidth}
    \caption{\textbf{Limitation with synthetic relighting.} The network is unable to extrapolate lighting variants for lighting characteristics that are not present in the original MID. The output is similar for both lighting probes that illuminate from beneath.
    } 
    \hfill
    \label{fig:relighting beneath}
  \end{minipage}
 \end{minipage}
 \vspace{-0.4cm}
\end{figure}

Inference through synthetic probes revealed two issues with the network. 
Firstly, it cannot generalize to angles of incidence that have not been present in the training dataset, such as illumination coming from beneath. Figure \ref{fig:relighting beneath} illustrates how 2 illumination sources generate the same flat image. Secondly, inference through synthetic probes requires the generation and the storage beforehand of each desired lighting condition. 

\begin{figure}[h]

    \centering
    \includegraphics[width=\linewidth]{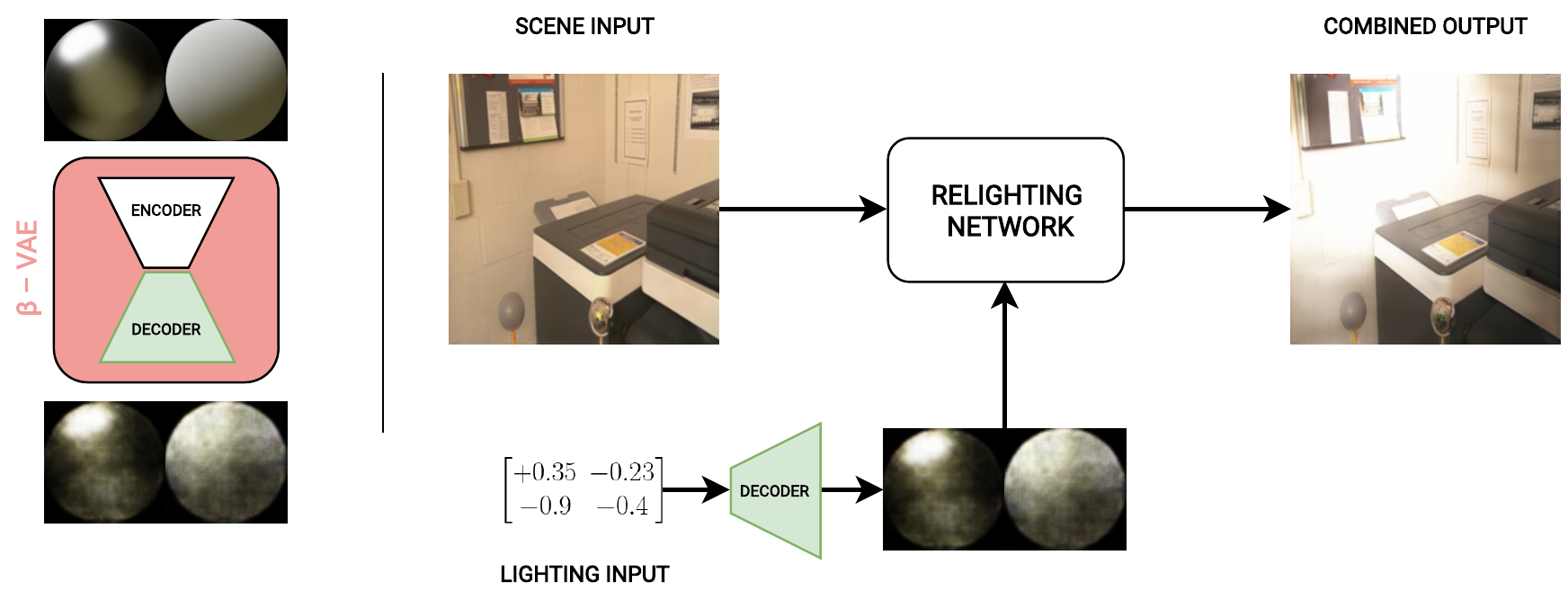}
    \caption{\textbf{Inference using the \textbeta{} - VAE decoder and latent space values.} \textbf{Left}: By training a \textbeta{} - VAE on the MID probes, it is possible to learn a representation of the illumination in the latent space. \textbf{Right}: By controlling the input tensor for the decoder, probes can be generated online for relighting inference.}
    \label{fig:VAE}
\end{figure}

Ideally, we would like to be able to directly influence  the lighting instead of relying on the existing probe images. VAEs are able to learn latent space representation for higher order information. However, disentangling features for complex images can be challenging, even more so when trying to separate lighting from geometry. We considered using a \textbeta{} - VAE \cite{beta-vae} to learn a latent space representation of simple probes, using solely the decoder to generate the probes online instead of relying on the existing generated synthetic probes (see Figure \ref{fig:VAE}). Although, we were able to achieve a limited control over the generated probes, lighting could not be properly isolated from the angle and the quality of reconstruction was lacking. Nonetheless, it seems that the actual quality of the probes did not influence the final relit image but only the relative contrasts within the probe. To ensure reproducibility and a more reliable control over the experiments we decided to use a mix of the averaged MID probes with additional Blender probes in the following benchmarks.

\begin{figure}[!h]
  \centering
   \hfill
  \begin{minipage}[h]{0.45\textwidth}
    \includegraphics[width=\textwidth]{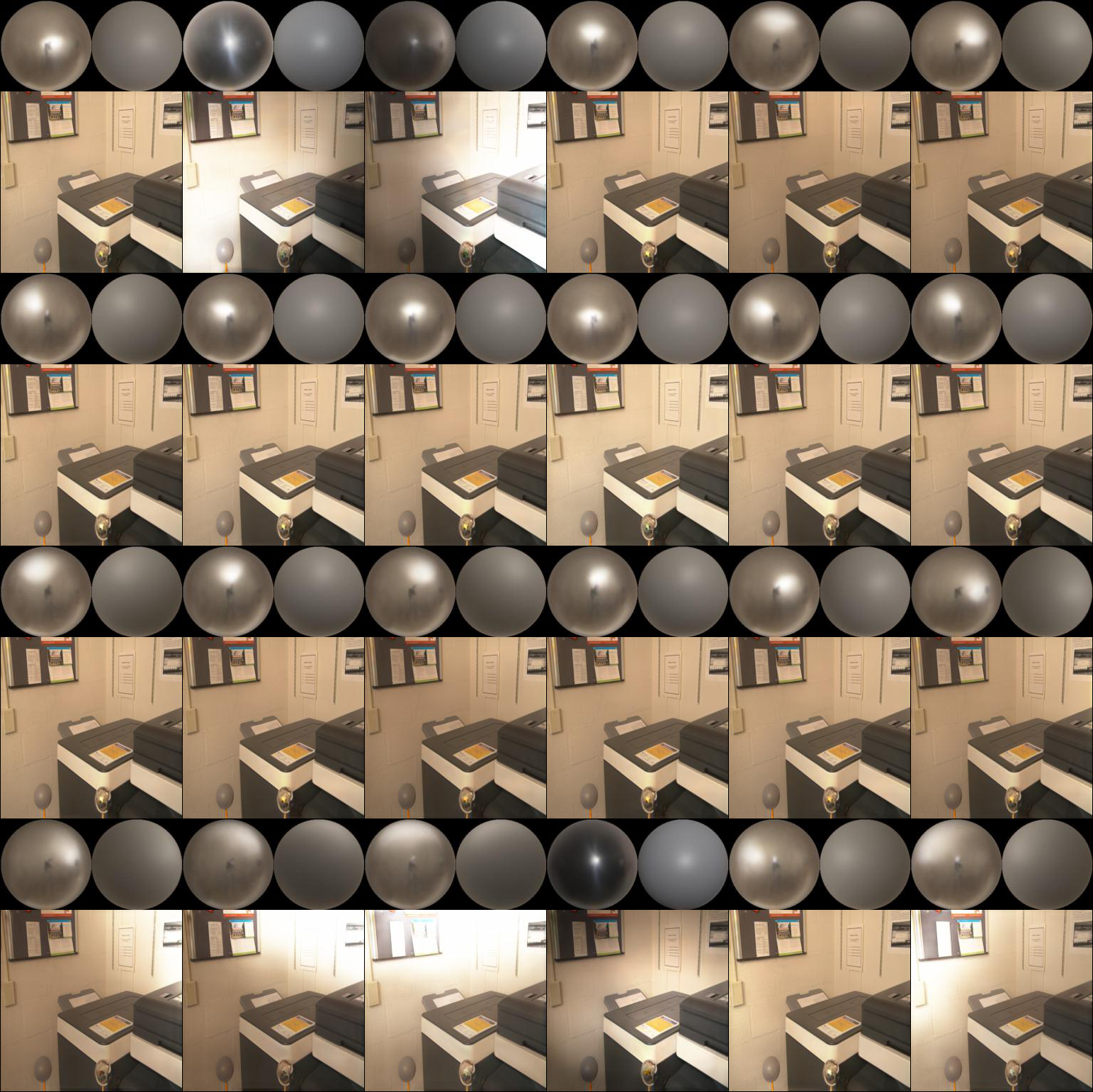}
  \end{minipage}
  \hfill
  \begin{minipage}[h]{0.45\textwidth}
    \includegraphics[width=\textwidth]{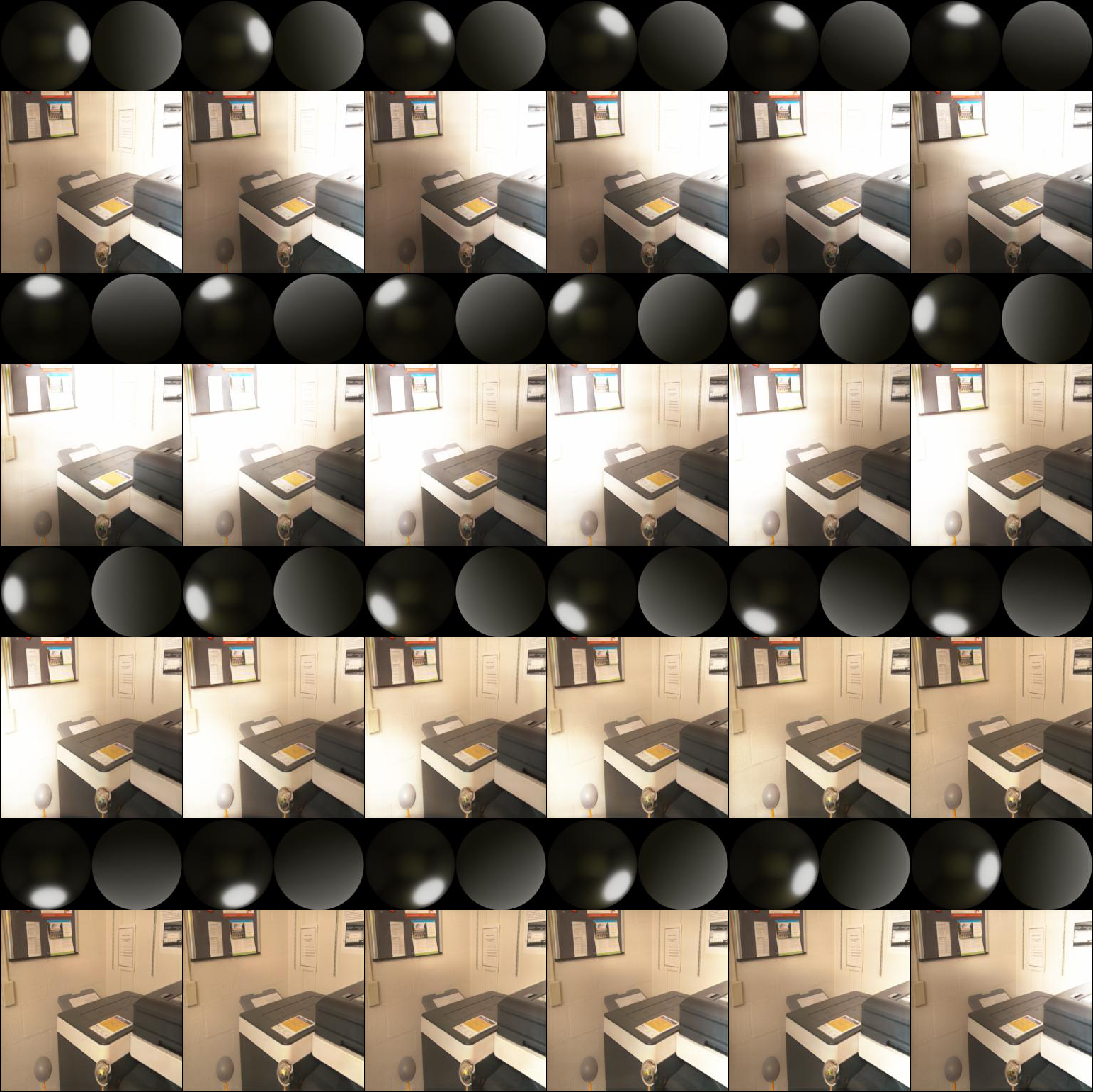}
  \end{minipage}
        \hfill
    \vspace{0.3cm}
    \caption{\textbf{Left}: 24 generated images using MID averaged probes as guides. \textbf{Right}: 24 generated images using synthetic Blender probes as guides.}
\label{fig:syntheticImages}
\vspace{-0.5cm}
\end{figure}

\vspace{-0.6cm}
\section{Benchmarks}
\vspace{-0.2cm}
\subsection{Experimental Setup}
\begin{figure}[h!]
\vspace{-0.4cm}
    \centering
    \includegraphics[width=0.9\linewidth]{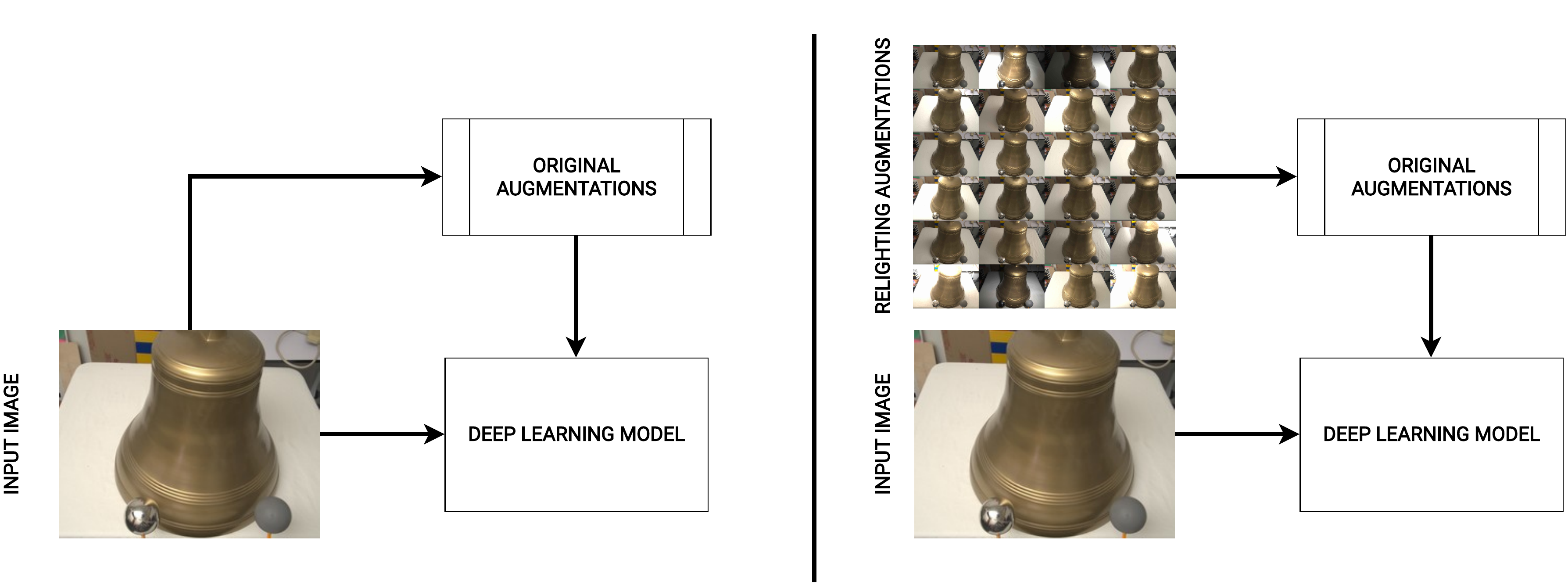}
    \vspace{-0.35cm}
    \caption{\textbf{Left}: Original pipeline as reported by the author using basic augmentations. \textbf{Right}: Augmented pipeline swapping the reference image with one of the relit variants before inputting it into the same basic augmentations.}
    \label{fig:training_setupl}
    \vspace{-0.5cm}
\end{figure}

\vspace{-0.2cm}
To explore the impact that our relighting network has, we decided to work with keypoint descriptors and detectors trained on various datasets. We establish a baseline for each model by training it with the original pipeline as described by their respective authors. For each image in the original training dataset, relit versions are generated and stored in memory. A new model is trained again with the same pipeline as the original. However, instead of feeding the original image in the transformations pipeline, a single version is randomly selected from a pool of the relit variants of that scene (see Figure \ref{fig:training_setupl}).
\vspace{-0.4 cm}
\subsection{HPatches}
\vspace{-0.2 cm}
HPatches~\cite{HPatches} evaluates the quality of image descriptors with respect to illumination and viewpoint changes. Specifically, we are interested in the Mean Matching Accuracy (MMA) as originally defined by \citet{r2d2}, which is the average percentage of correct matches in an image pair at varying thresholds of pixel error. Evaluating R2D2~\cite{r2d2} on this benchmark demonstrates clear results on the potential that relighting augmentation can have on descriptor quality. When augmenting the COCO~\cite{COCO}, VIDIT~\cite{vidit} and Aachen~\cite{VisualLoc} datasets through relighting, R2D2 consistently sees an improvement in all metrics as seen in Table \ref{tab:R2D2 HPatches} with illumination scores seeing the highest improvements. We compared our results (dubbed "Deep Relighting") to the same pipeline augmented with standard illumination augmentation using the Albumentations library \cite{albumentation} and style transfer augmentation using FastPhotoStyle \cite{li2018closedform}, similarly to \citet{melekhov2020image}. In all cases we observed similar or better results while being up to 6 times faster than FastPhotoStyle (0.14 seconds vs 0.91 seconds on average). Interestingly, we noticed that in the case of small datasets, R2D2 was not able to detect keypoints if the training data presented too much variation between pairs, such as with augmentations provided through FastPhotoStyle. \\
Furthermore, by simply augmenting the Aachen dataset through our pipeline, we were able to achieve the same best result as the authors reported in their paper while the original R2D2 model was trained on a combination of four datasets as referenced in Table \ref{tab:R2D2 HPatches Best}. 

\begin{table}[h]
	\footnotesize
	\centering
		\begin{tabular}{llll}
			\hline
			Training Dataset & Overall & Viewpoint & Illumination\\
			\hline
			\hline
            COCO - Base             & 70.0  & 64.7    & 75.6       \\
            COCO - Albumentations \cite{albumentation}       & 71.1  & 64.8    & 77.9       \\
            COCO - FastPhotoStyle \cite{li2018closedform}       & 72.7  & 65.9   & \textbf{80.0}       \\
            COCO - Deep Relighting        & \textbf{73.3}   & \textbf{67.0}  & \textbf{80.0}    \\
            \hline
            VIDIT - Base                  & 66.7  & 73.6    & \textbf{60.2}      \\
            VIDIT - Albumentations \cite{albumentation}       & 68.0  & 78.1   & 58.65     \\
            VIDIT - FastPhotoStyle \cite{li2018closedform}       & *  & *   & *      \\
            VIDIT - Deep Relighting        & \textbf{68.9}   & \textbf{78.2}  & \textbf{60.2}    \\

			\hline
        \end{tabular}
        
	\normalsize
	\vspace{0.4 cm}
    \caption{\textbf{Mean Matching accuracy of different variants of R2D2 on HPatches~\cite{HPatches}.} We consider two training sets: COCO~\cite{COCO} and VIDIT~\cite{vidit}, and use an error threshold of 3 pixels. "*" indicates a failure to detect keypoints. Our relighting augmentation consistently outperforms the baseline augmentation techniques.}
	\label{tab:R2D2 HPatches}
	\vspace{-0.3cm}
\end{table}

\begin{table}[h]
	\footnotesize
	\centering
		\begin{tabular}{llll}
			\hline
			
			Training Dataset & Overall & Viewpoint & Illumination\\
			\hline
						\hline

            Aachen - Base                    & 69.4  & 64.1     & 75.0       \\
            Aachen - WAF                  & 70.9  & \textbf{66.5}    & 75.6       \\
            Aachen - WASF                 & 71.6  & 66.1    & 77.4       \\
            Aachen - Deep Relighting                    & \textbf{72.6}  & 66.0   & \textbf{79.6}      \\
                           \hline
            Aachen - WASF - MS                & \textbf{75.8}  & \textbf{71.1}    & 80.8       \\
            Aachen - Deep Relighting - MS                & \textbf{75.8}  & 70.0    & \textbf{82.0}      \\

			\hline
        \end{tabular}
        
	\normalsize
	\vspace{0.4 cm}
    \caption{\textbf{Effects of relighting augmentation for small datasets.} We compare the MMA at 3 pixels error threshold on HPatches~\cite{HPatches} of R2D2 trained only on the Aachen dataset~\cite{VisualLoc} augmented by our method, or trained on the original datasets used by \citet{r2d2} (W: Web images, A: Aachen dataset, S: Aachen with style transfer, F: Aachen with optical flow, MS: multi-scale extraction).}
	\label{tab:R2D2 HPatches Best}
	\vspace{-0.5cm}
\end{table}

\vspace{-0.1 cm}
\subsection{RDNIM}
\vspace{-0.1 cm}
The RDNIM \cite{RDNIM} benchmark builds on top of the Day-Night Image Matching \cite{DNIM} benchmark by adding rotation warps. RDNIM measures a homography estimation score by computing the distance between the image from the true homography warping and the image warped with the estimated one, as described in \cite{superpoint}. The precision and the recall are respectively the ratio of correct matches over all matches and the number of correct matches over true matches. All metrics are using a distance threshold of 3 pixels to consider a match as correct. Although our model should only be able to affect invariance with regards to illumination, we decided to test our model behaviour on this benchmark as it also contains day-night matching.

We, once again, compare  our relighting against previous augmentation techniques in Table~\ref{tab:rdnim_baselines} and show improved homography estimation and competitive precision-recall. The lower improvements compared to HPatches might be explained by the fact that our relighting network was trained on the indoor scenes of the MID, while the RDNIM dataset consists of outdoor day-night image pairs. We additionally show that our augmentation can generalize to other models, such as SuperPoint~\cite{superpoint} in Table~\ref{tab:rdnim_methods}. Our augmentation is still able to increase the performance of SuperPoint, but with a smaller added value due to the fact that SuperPoint already benefits from more advanced photometric augmentation techniques than R2D2.

\begin{table}[h]
	\footnotesize
	\centering
		\begin{tabular}{lllll}
			\hline
            Reference &Training Dataset & H. Estimation & Precision & Recall  \\
			\hline
			\hline

\hline

\multirow{4}{*}{Day} & COCO - Base                                & 13.9                  & 20.3      & 3.8    \\
&COCO - Albumentations \cite{albumentation}                     & 13.9                  & 21.3      & 3.8    \\
&COCO - FastPhotoStyle  \cite{li2018closedform}                    & 15.2                  & \textbf{26.1}      & \textbf{4.3}    \\
&COCO - Deep Relighting                     & \textbf{15.4}                  & 25.3      & 3.9  \\
\hline
\multirow{4}{*}{Night} & COCO - Base                                & 17.9                  & 25.3      & 0.5    \\
&COCO - Albumentations \cite{albumentation}                     & 18.1                  & 26.6      & 0.5    \\
&COCO - FastPhotoStyle \cite{li2018closedform}                    & 18.5                  & \textbf{31.3}      & \textbf{0.6}    \\
&COCO - Deep Relighting                     & \textbf{19.4}                  & 30.0      & 0.5  \\
			\hline
        \end{tabular}
	\vspace{0.2 cm}
    \caption{\textbf{Matching metrics of R2D2 on RDNIM~\cite{RDNIM} under different augmentations.}}
	\label{tab:rdnim_baselines}
\end{table}
\vspace{-0.2 cm}

\begin{table}[h]
	\footnotesize
	\centering
    \setlength{\tabcolsep}{1mm}
        \begin{tabular}{llllll}
			\hline
            Model & Reference &Training Dataset & H. Estimation & Precision & Recall  \\
			\hline
			\hline
\multirow{4}{*}{R2D2 \cite{r2d2}} & \multirow{2}{*}{Day} & COCO - Base                                & 13.9                  & 20.3      & 3.8    \\
& & COCO - Deep Relighting                     & \textbf{15.4}  \textcolor{darkgreen}{(+11\%)}                & \textbf{25.3}  \textcolor{darkgreen}{(+25\%)}     & \textbf{3.9} \textcolor{darkgreen}{(+2.6\%)} \\
\cmidrule(lr){2-6}
& \multirow{2}{*}{Night} & COCO - Base                                & 17.9                  & 25.3      & \textbf{0.5}    \\
& & COCO - Deep Relighting                     & \textbf{19.4}   \textcolor{darkgreen}{(+8.4\%)}               & \textbf{30.0}   \textcolor{darkgreen}{(+19\%)}   & \textbf{0.5} \textcolor{black}{(0.0\%)} \\
\hline
\multirow{4}{*}{SuperPoint \cite{superpoint}} & \multirow{2}{*}{Day}	& COCO - Base               & 11.7                  & 17.0      & 10.3   \\
& & COCO - Deep Relighting           & \textbf{12.2} \textcolor{darkgreen}{(+4.3\%)}  & \textbf{17.3}  \textcolor{darkgreen}{(+1.8\%)}    & \textbf{11.2} \textcolor{darkgreen}{(+8.7\%)}   \\
\cmidrule(lr){2-6}
& \multirow{2}{*}{Night}	& COCO - Base               & \textbf{17.5}                  & 26.7      & 16.7   \\
& & COCO - Deep Relighting           & 16.3 \textcolor{red}{(-6.8\%)} & \textbf{26.8} \textcolor{darkgreen}{(+0.04\%)}     & \textbf{17.6} \textcolor{darkgreen}{(+5.3\%)}  \\
			\hline
        \end{tabular}
        
	\vspace{0.2 cm}
    \caption{\textbf{Generalization of our augmentation scheme to multiple models.}}
	\label{tab:rdnim_methods}
\end{table}
\vspace{-0.3 cm}
\subsection{Visual Localization: Aachen Day-Night}
\vspace{-0.1 cm}
The Long-Term Visual Localization benchmark~\cite{VisualLoc} measures the capacity of a model to estimate the position of a camera with respect to a reference query image within various degrees of precision. We are interested in the local feature challenge v1.1 of the Aachen Day-Night dataset.
We have decided to test how our pipeline performs with the two models SuperPoint~\cite{superpoint} and R2D2~\cite{r2d2}. We saw significant improvements in their respective performances for both training datasets COCO and VIDIT at all tolerance levels (see Table \ref{tab:R2D2 Visuallocalisation}). It is worth noting that VIDIT underwent the largest improvement. This is probably due to the very limited base training dataset consisting of only 300 scenes, highlighting the importance of data augmentation when data is scarce.

\begin{table}[h]
\footnotesize
    \vspace{-0.1 cm}
	\centering
		\begin{tabular}{lllll}
			\hline
		Model & Training Dataset              & 0.25m, 2° & 0.5m, 5°  & 5m, 10°  \\
			\hline
			\hline

        \multirow{4}{*}{R2D2 \cite{r2d2}} & COCO - Base                                      & 60.2              & 74.3               & 86.4            \\
        & COCO - Deep Relighting                           & \textbf{63.9} \textcolor{darkgreen}{(+6.1\%)}             & \textbf{78.5} \textcolor{darkgreen}{(+5.6\%)}              & \textbf{92.7}  \textcolor{darkgreen}{(+7.3\%)}          \\

       & VIDIT - Base         & 40.8              & 53.4             & 61.3                     \\
        & VIDIT - Deep Relighting      & \textbf{53.9}   \textcolor{darkgreen}{(+32\%)}           & \textbf{71.2}  \textcolor{darkgreen}{(+33\%)}           & \textbf{83.2} \textcolor{darkgreen}{(+36\%)}                   \\

  \hline
  \multirow{4}{*}{SuperPoint \cite{superpoint}}  & COCO - Base                   & 58.6              & 69.6             & 81.7            \\
        & COCO - Deep Relighting              & \textbf{62.3} \textcolor{darkgreen}{(+6.3\%)}             & \textbf{74.9} \textcolor{darkgreen}{(+7.6\%)}            & \textbf{86.4}  \textcolor{darkgreen}{(+5.7\%)}          \\
        &VIDIT - Base                   & 49.2              & 63.4             & 72.3            \\
        &VIDIT - Deep Relighting              & \textbf{54.5} \textcolor{darkgreen}{(+11\%)}             & \textbf{68.1}  \textcolor{darkgreen}{(+7.4\%)}           & \textbf{78.5}   \textcolor{darkgreen}{(+8.5\%)}         \\

        \hline
        \end{tabular}
        
	\normalsize
	\vspace{0.1 cm}
        \caption{\textbf{Results on the Visual Localization challenge for "Aachen v1.1" at night}}
	\label{tab:R2D2 Visuallocalisation}
\end{table}
\vspace{-0.2 cm}

\vspace{-0.4cm}
\section{Conclusion}
\vspace{-0.2 cm}
We presented a deep learning relighting tool that is able to quickly generate multiple illumination variants when given an input scene and user parameters. We demonstrated, on several benchmarks, that it performs at least as well or better than existing augmentation techniques and style transfer based relighting. Furthermore, models trained with our pipeline seem to have the most drastic improvements on datasets that have fewer original samples. However, while our network serves its purpose as a data augmentation tool, the photo-realism can still be improved and the model could be extended to better accommodate outdoor scenes. 
\clearpage
\bibliography{egbib.bib}
\end{document}